\documentclass[twocolumn,10pt]{article}

\usepackage[utf8]{inputenc}
\usepackage[T1]{fontenc}
\usepackage{amsmath,amssymb,amsfonts}
\usepackage{graphicx}
\usepackage{hyperref}
\usepackage[numbers,sort&compress]{natbib}
\usepackage{algorithm}
\usepackage{algorithmic}
\usepackage{booktabs}
\usepackage{caption}
\usepackage{subcaption}
\usepackage[margin=1in]{geometry}

\hypersetup{
    colorlinks=true,
    linkcolor=blue,
    citecolor=blue,
    urlcolor=blue
}

\title{
    \vspace{-1cm}
    \rule{\textwidth}{1pt}
    \vspace{0.3cm}
    \Large\textbf{Humanlike Multi-user Agent (HUMA): Designing a Deceptively Human AI Facilitator for Group Chats}
    \vspace{0.3cm}
    \rule{\textwidth}{1pt}
}
\author{
    \textbf{Mateusz Jacniacki}\\
    Soofte Research \\
    \texttt{mjacniacki@soofte.com} \\
    \and
    \textbf{Marti Carmona Serrat} \\
    Soofte Research\\
    \texttt{mtk@soofte.com}
}
\date{\today}

\begin{document}

\maketitle

\raggedbottom

\begin{abstract}
Conversational agents built on large language models (LLMs) are becoming increasingly prevalent, yet most systems are designed for one-on-one, turn-based exchanges rather than natural, asynchronous group chats. As AI assistants become widespread throughout digital platforms, from virtual assistants to customer service, developing natural and humanlike interaction patterns seems crucial for maintaining user trust and engagement. We present the Humanlike Multi-user Agent (HUMA), an LLM-based facilitator that participates in multi-party conversations using human-like strategies and timing. HUMA extends prior multi-user chatbot work with an event-driven architecture that handles messages, replies, reactions and introduces realistic response-time simulation. HUMA comprises three components—Router, Action Agent, and Reflection—which together adapt LLMs to group conversation dynamics.

We evaluate HUMA in a controlled study with 97 participants in four-person role-play chats, comparing AI and human community managers (CMs). Participants classified CMs as human at near-chance rates in both conditions, indicating they could not reliably distinguish HUMA agents from humans. Subjective experience was comparable across conditions: community-manager effectiveness, social presence, and engagement/satisfaction differed only modestly with small effect sizes. Our results suggest that, in natural group chat settings, an AI facilitator can match human quality while remaining difficult to identify as nonhuman. 
\end{abstract}

\section{Introduction}
\label{sec:introduction}

Since 2023 we have seen a surge in development and adoption of LLM-based virtual assistants across diverse digital platforms. Despite their growing capabilities, these systems often fail to feel genuinely human in their interactions. As AI agents become more human-like, they appear to risk falling into an ``uncanny valley'' where near-human but imperfect behavior triggers discomfort \cite{ciechanowski2019uncanny}. When chatbots respond with superhuman speed or maintain unnaturally consistent engagement, users may experience a sense of disconnect \cite{gnewuch2022opposing}.

One domain where human-like AI interaction appears particularly critical is community management and facilitation. Community platforms with group chat interfaces continue to grow in popularity; for example, Discord reached 200 million monthly active users in 2025. Yet how AI assistants should participate in and facilitate group conversations remains underexplored relative to one-on-one settings. This domain presents unique challenges and serves as an ideal testbed for evaluating human-like AI behavior in natural, multi-party interactions.

This paper presents the Humanlike Multi-user Agent (HUMA), an LLM-based facilitator designed to follow natural interaction patterns of human participants in online group chats. HUMA addresses three core challenges intrinsic to multi-party, asynchronous conversation: deciding when to speak versus stay silent, determining whom to address and how, and managing interruptions in the presence of rapid, overlapping messages. To support believable behavior, HUMA simulates human response timing and typing dynamics.

We evaluate HUMA in live, four-person role-play chats comparing an AI facilitator to human community managers. Two questions guide our study: (RQ1) Can participants reliably distinguish HUMA from human facilitators? and (RQ2) Do participants report comparable satisfaction, engagement, and social presence when interacting with HUMA versus human CMs? In our experiment with 97 participants sourced from Prolific, detection rates hovered near chance, and subjective experience ratings were similar across conditions with only small differences.

Our contributions are threefold: (1) a framework for event-driven, human-like group chat agent that integrates diverse conversational patterns, timing simulation, interruption handling, and reflection; (2) a set of conversational strategies and tooling constraints tailored for natural group chat dynamics; and (3) a controlled human-subject evaluation demonstrating that an AI facilitator can be both difficult to identify as nonhuman and comparable to human CMs on key experience measures.

\section{Related Work}
\label{sec:related}

Our work builds on research in LLM-based conversational agents, multi-party dialogue systems, and human-like behavior simulation.

\textbf{LLM-based Conversational Agents.} Contemporary LLM-based agents have demonstrated remarkable capabilities in simulating human-like behavior across diverse contexts \cite{park2023generative,zhao2025agentsociety,wang2024llmagent}, including matching human performance in social situational judgments \cite{zhao2024llmsocial}. However, most existing research focuses on one-on-one interactions where the agent determines only ``What'' to respond, following predictable turn-taking structures \cite{shuster2023multiparty,addlesee2023designing,weizenbaum1966eliza}. Our work extends this foundation by addressing the additional challenges that emerge when multiple human participants interact simultaneously with an AI facilitator.

\textbf{Multi-party Dialogue Systems.} The MUCA framework \cite{mao2024muca} formalized the ``3W'' design dimensions for multi-user chatbots—``What'' to say, ``When'' to respond, and ``Who'' to answer—introducing conversational strategies for goal-oriented group discussions. Recent work has explored proactive agents with ``inner thoughts'' mechanisms \cite{zhang2025proactive} and multi-agent frameworks for collaborative conversations \cite{chen2024captain,wu2023autogen}. However, existing systems primarily focus on goal-oriented conversations with explicit turn-taking, whereas our work addresses more natural, asynchronous group chat environments where participation is optional.

\textbf{Human-like Behavior Simulation.} Achieving human-like behavior appears to require attention to both linguistic patterns and temporal dynamics, with research suggesting agents can replicate human behavior in various contexts \cite{park2023generative,wang2024recagent,zhang2024generative}. Critical to believability appears to be simulating realistic response timing—human typing speeds of 50–100 WPM versus rapid LLM generation can create an ``uncanny valley'' effect \cite{typing2023average,tenbosch2005pauses,replika2023study}. Our work specifically addresses temporal dynamics including typing indicators, message delivery patterns, and vulnerability to interruption—elements that appear essential for natural group conversation but are absent from many existing frameworks.

\textbf{Our Contribution.} HUMA integrates multi-party dialogue design with human-like timing and event-driven interaction to enable natural participation in group chats. We extend MUCA's 3W framework with an architecture that supports asynchronous messaging, reactions, replies, and interruption-aware behavior. Our evaluation focuses on the agent's ability to pass as human in natural group conversations and to deliver subjective experiences comparable to those provided by human facilitators.

\section{Framework Architecture}
\label{sec:methodology}

\subsection{Challenges of Group Chat Environments}
\label{subsec:group-chat-challenges}

Contemporary LLMs and conversational agents are predominantly designed for two-party dialogue \cite{yi2024surveyrecentadvancesllmbased,shuster2023multiparty}. However, multi-party conversational settings introduce new challenges that appear to require rethinking core assumptions about conversational AI \cite{li2022whomsays,mahajan2025multiparty}. In contrast to one-on-one exchanges, group conversations exhibit distinct characteristics that complicate agent design:

\textbf{Optional participation.} Group conversations often proceed between participants without requiring agent involvement. Participants communicate with each other, and not every exchange necessitates contribution from all group members. The agent must recognize when its participation appears appropriate and when to stay silent.

\textbf{Asynchronous messaging.} Participants frequently send multiple consecutive messages rather than waiting for responses after every utterance. A single thought expressed over several messages might be interrupted by other participants, making it harder to keep up with conversation flow.

\textbf{Dynamic conversation flow.} As participants formulate their thoughts, new messages may arrive, potentially invalidating current reasoning. A human typing a reply might abandon or modify their message upon seeing new information; similarly, when participants send messages rapidly, the agent cannot afford to spend long processing each message individually. The agent must exhibit similar adaptability and consider the evolving conversation state.

\textbf{Diverse interaction forms.} Beyond text messages, participants use reaction emojis and replies, requiring the agent to understand and employ these communication primitives appropriately. In settings where typing indicators and message delivery status are present, simply interacting with the platform can itself communicate intent (e.g., starting to type signals that a response is forthcoming).

Achieving human-like behavior in group conversations appears to require two additional design constraints beyond handling the challenges above: (1) adopting linguistic patterns and tone appropriate for informal group settings, and (2) matching human response timing patterns. The latter appears particularly critical — by simulating realistic time delays required to compose a message, the agent becomes vulnerable to interruptions in ways that mirror human experience.

\subsection{Event-Driven Architecture Overview}
\label{subsec:architecture-overview}

Upon joining a conversation, HUMA receives the complete chat history and list of participants. Subsequently, it processes discrete events in real time as they occur:
\begin{itemize}
\item Participant joins the chat
\item Message is sent
\item Reaction is added
\item Reaction is removed
\item Reply to a message is sent
\item Typing indicator is activated
\end{itemize}

\begin{figure*}[!t]
\centering
\includegraphics[width=0.8\textwidth]{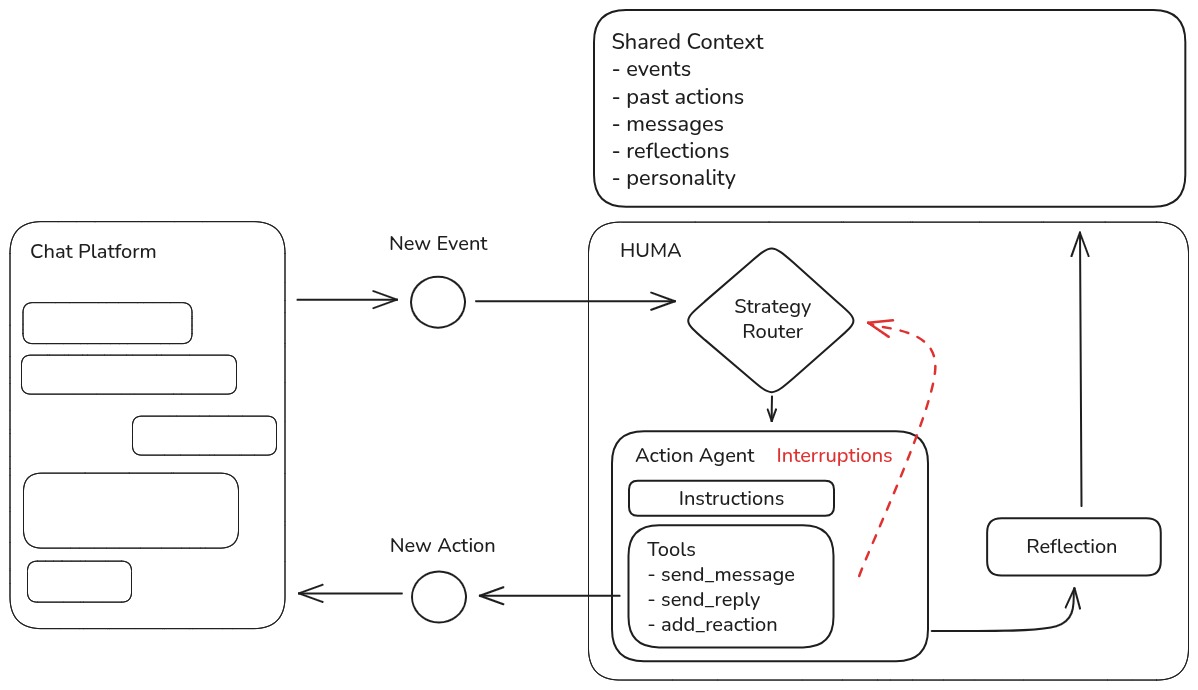}
\caption{HUMA system architecture. Events from the chat platform trigger a three-stage workflow: the Router selects an appropriate conversational strategy, the Action Agent executes it using available tools, and the Reflection component synthesizes context for future iterations. The workflow can be interrupted by new events, enabling natural adaptation to rapid conversation dynamics.}
\label{fig:architecture}
\end{figure*}

Each event triggers a three-stage workflow: routing, action execution, and reflection. Crucially, this workflow can be interrupted mid-execution by incoming events, forcing the agent to reassess its intended actions while preserving its internal scratchpad. Figure~\ref{fig:architecture} illustrates the overall system architecture and event flow.

\subsection{Router: Strategy Selection}
\label{subsec:router}

The router addresses both the ``When'' and ``What'' of agent participation by selecting from 20 predefined \textit{conversational strategies} inspired by the MUCA framework \cite{mao2024muca}. These include ``Keep Silent,'' ``Go Deeper,'' ``Ask Question,'' ``Bridge Perspectives,'' ``Recall Message,'' ``Refocus to Goal,'' and others chosen to facilitate diverse, context-sensitive responses.

Strategy selection combines two scoring dimensions:
\begin{itemize}
\item \textbf{Appropriateness} ($A \in [0,1]$): An LLM evaluates each strategy's contextual fit given the full conversation history and strategy descriptions, returning a score for each strategy.

\item \textbf{Timeliness} ($T \in [0,1]$): Promotes behavioral diversity by penalizing recently-used strategies. The system tracks the last $N$ strategy activations, where $N$ equals the total number of available strategies. For a
strategy last used $k$ steps ago in this history, the timeliness score is:
$$T_s = \min\left(1, \frac{k}{N}\right)$$
A just-used strategy has $T=0$ and gradually recovers as other strategies are used, gaining $\frac{1}{N}$ per step until reaching $T=1$ after $N$ different strategy uses.
Several strategies are exceptions from Timeliness score and maintain $T_s = 1$ regardless of their usage frequency. These include ``Keep Silent,'' ``Directly Mentioned,'' ``Continue Pending,'' and ``Tell a Story.''

\end{itemize}

The router selects the strategy maximizing the combined score $A + T$. Certain strategies are exempt from timeliness penalties and maintain $T = 1$ regardless of usage frequency—for example, ``Keep Silent'' (to allow natural pauses) and ``Tell a Story'' (to support multi-message narratives).

The router outputs a single conversational strategy with the highest combined score, which is then passed to the action agent for execution.

\subsection{Action Agent: Strategy Execution}
\label{subsec:action-agent}

Given the selected strategy, the action agent executes it using three tools: \texttt{send\_message}, \texttt{send\_reply}, and \texttt{add\_reaction}. The agent maintains a scratchpad for intermediate reasoning that persists across potential interruptions.

\subsubsection{Delay in Responses}

Natural human typing speed is typically around 50–100 WPM, whereas most non-reasoning commercially available (through API) LLMs generate tokens much faster. Introducing delay to simulate typing speed therefore appears necessary. This is implemented within the \texttt{send\_message} and \texttt{send\_reply} tools' execution logic. During simulated typing the agent remains susceptible to interruptions.

\subsubsection{Interrupt Handling}

Timing simulation makes HUMA vulnerable to interruptions in ways that mirror human experience. Interrupts arriving during the LLM generation phase are queued until generation completes. When the agent receives an interrupt, the workflow restarts from the Router component. Crucially, the scratchpad and intended tool calls are preserved and included in the new context, allowing the agent to be aware of its interrupted intentions. One of the available conversational strategies, ``Continue Pending Action,'' specifically allows the agent to resume interrupted activities.

\subsubsection{Tool Call Constraints}

The system supports parallel tool calls, enabling the agent to simultaneously send messages and add reactions. However, parallel \texttt{send\_message} calls are explicitly forbidden; attempting multiple message sends in a single turn results in an error response for the second and subsequent attempts. We guide the LLM in its system prompt to prefer multi-turn, sequential \texttt{send\_message} tool calls instead.

\subsection{Reflection: Maintaining Coherence}
\label{subsec:reflection}

Following uninterrupted action execution, HUMA generates a single-sentence reflection synthesizing the conversational context and its recent behavior. This reflection serves two purposes: exploring potential conversational topics adjacent to the current one and maintaining a coherent conversation arc across multiple workflow runs. The reflection is then included in the agent's context for subsequent iterations.

\section{Evaluation}
\label{sec:results}

We evaluate HUMA's performance through a controlled human-subject experiment examining two critical dimensions of group chat facilitation. Our evaluation addresses whether AI agents can achieve functional parity with human facilitators in natural group conversations.

Our investigation centers on two research questions:

\textbf{RQ1 (Human-likeness):} Can participants reliably distinguish HUMA from human community managers in group chat settings? If HUMA successfully implements human-like behavioral patterns, it should be indistinguishable from human facilitators.

\textbf{RQ2 (Subjective Experience):} Do participants report comparable satisfaction, engagement, and social presence when interacting with HUMA versus human community managers?

\subsection{Experimental Design}

We conducted a between-subjects study comparing AI and human community managers in role-playing group chats. Participants were randomly assigned to one of two conditions: Control (human community manager) or Treatment (AI community manager). All groups contained four participants—one community manager and three regular users.

\subsubsection{Roles and Scenario}

Participants joined a simulated online community for Leonardo AI users, a generative AI art platform. Each participant received one of four roles with distinct personas and goals: The Community Manager (facilitator managing the group), The Interested (newcomer seeking help with specific creative projects), The Regular (experienced user sharing tips and maintaining conversation), and The Critic (skeptical experienced user questioning claims and demanding evidence). This scenario mirrors authentic online community dynamics where members have different experience levels, motivations, and communication styles.

\subsubsection{Procedure}

The complete session lasted around 30 minutes, including preparation, waitroom, live chat, survey, and debriefing phases. The live group chat portion lasted approximately 10 minutes within the session.

\subsubsection{Role Induction}

Participants completed three preparation screens to establish their character's perspective and give them authentic conversational material. First, they crafted a detailed backstory by answering role-specific prompts (e.g., Community Managers described their most viral artwork and stated their community values; Interested members specified what art they wanted to create and what blocked them; Regular users shared their favorite techniques). Second, they completed a 100+ character writing exercise simulating realistic community interactions (e.g., explaining how to fix a common technical issue, describing a project goal). Third, they answered reflection questions about their character's goals, emotions, and current state of mind. Finally, they selected nicknames. This multi-stage preparation provided topics to discuss, personal stakes, and emotional grounding to simulate genuine community behavior.

\subsubsection{Platform}

The chat interface supported real-time messaging, message replies, emoji reactions, and typing indicators. A countdown timer displayed remaining time. Both human and AI community managers received identical facilitation instructions emphasizing welcoming members, sharing technical advice, maintaining positive discussions, and encouraging participation.

\subsubsection{Measures}

Immediately following chat, participants completed a 25-item survey across six pages measuring: community manager effectiveness (7 items, 1–5 Likert), social presence (3 items), engagement and satisfaction (3 items), human-likeness and competence (2 items), an attention check, open-ended coaching feedback to the community manager, and AI detection questions. The AI detection block asked participants to classify the community manager as human or AI, rate confidence (1–7 scale), explain their reasoning (50+ characters), and report familiarity with AI chatbots (1–5 scale). Placing detection questions last minimized priming effects on earlier satisfaction and quality ratings.

\subsection{Participants}

We analyzed data from 97 participants recruited through Prolific (English-fluent, aged 18–40, with prior platform experience) across two study conditions (41 in the Human CM condition, 56 in the AI CM condition). All participants completed the post-chat survey immediately following their chat session.

\subsection{Results: AI Detection (RQ1)}

Participants were unable to distinguish AI community managers from human ones. In the AI CM condition (n=56), participants identified the community manager as AI 55.4\% of the time (95\% CI: [42.4\%, 67.6\%]). In the Human CM condition (n=41), participants identified the community manager as human only 46.7\% of the time (95\% CI: [30.2\%, 63.9\%]).

\begin{figure}[t]
\centering
\includegraphics[width=\columnwidth]{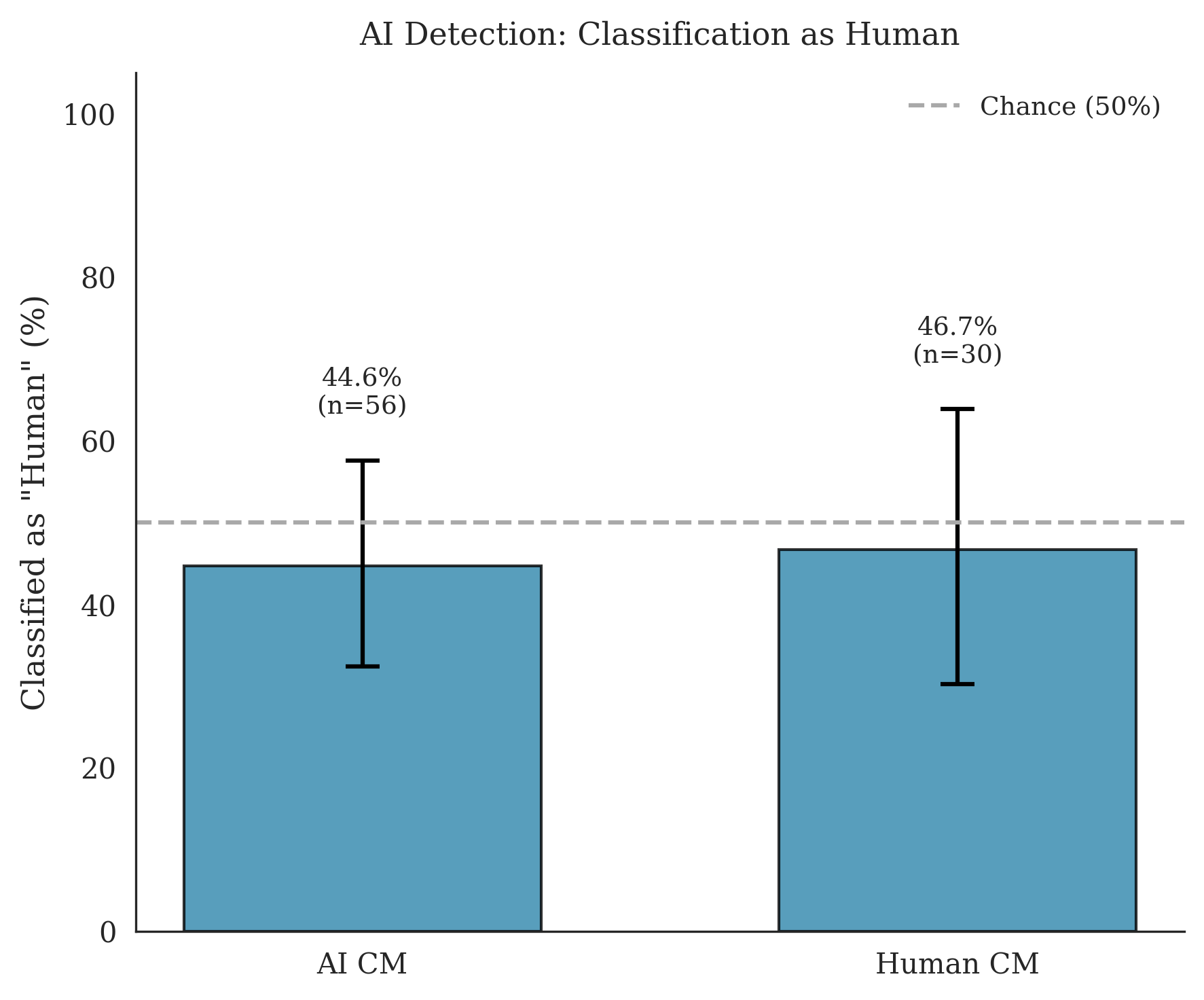}
\caption{AI detection results showing the percentage of participants classifying the community manager as ``human'' in each condition, with 95\% Wilson score confidence intervals. Both conditions cluster near chance level (50\%, dashed line). Participants exhibited symmetric confusion, unable to distinguish AI from human community managers.}
\label{fig:ai-detection}
\end{figure}

Classification rates were nearly identical across conditions. Participants classified community managers as ``human'' at similar rates whether interacting with actual AI CMs (44.6\%) or actual human CMs (46.7\%). Figure~\ref{fig:ai-detection} illustrates this symmetric confusion, with both conditions showing overlapping confidence intervals centered near 50\%, indicating classification rates did not differ meaningfully from chance.

Detection confidence averaged 3.91 out of 7 for AI CMs and 4.87 out of 7 for human CMs, indicating moderate uncertainty in both cases. These patterns suggest participants lacked reliable cues to distinguish AI from human facilitators.

\subsection{Results: Conversation Quality and Satisfaction (RQ2)}

Survey responses revealed similar experiences across conditions (Table~\ref{tab:scales}). Community Manager Effectiveness scores averaged 4.48 (SD=1.06) for human CMs and 4.14 (SD=0.84) for AI CMs, a difference of 0.34 points on a 5-point scale (Cohen's $d = -0.37$). Social Presence scores were 4.89 (SD=0.87) for human CMs versus 4.71 (SD=0.83) for AI CMs ($d = -0.21$). Engagement \& Satisfaction showed means of 4.58 (SD=0.97) and 4.32 (SD=0.78) respectively ($d = -0.30$). Human-likeness \& Competence demonstrated nearly identical scores: 4.13 (SD=1.30) for human CMs and 4.14 (SD=0.92) for AI CMs ($d = 0.01$).

\begin{figure}[t]
\centering
\includegraphics[width=\columnwidth]{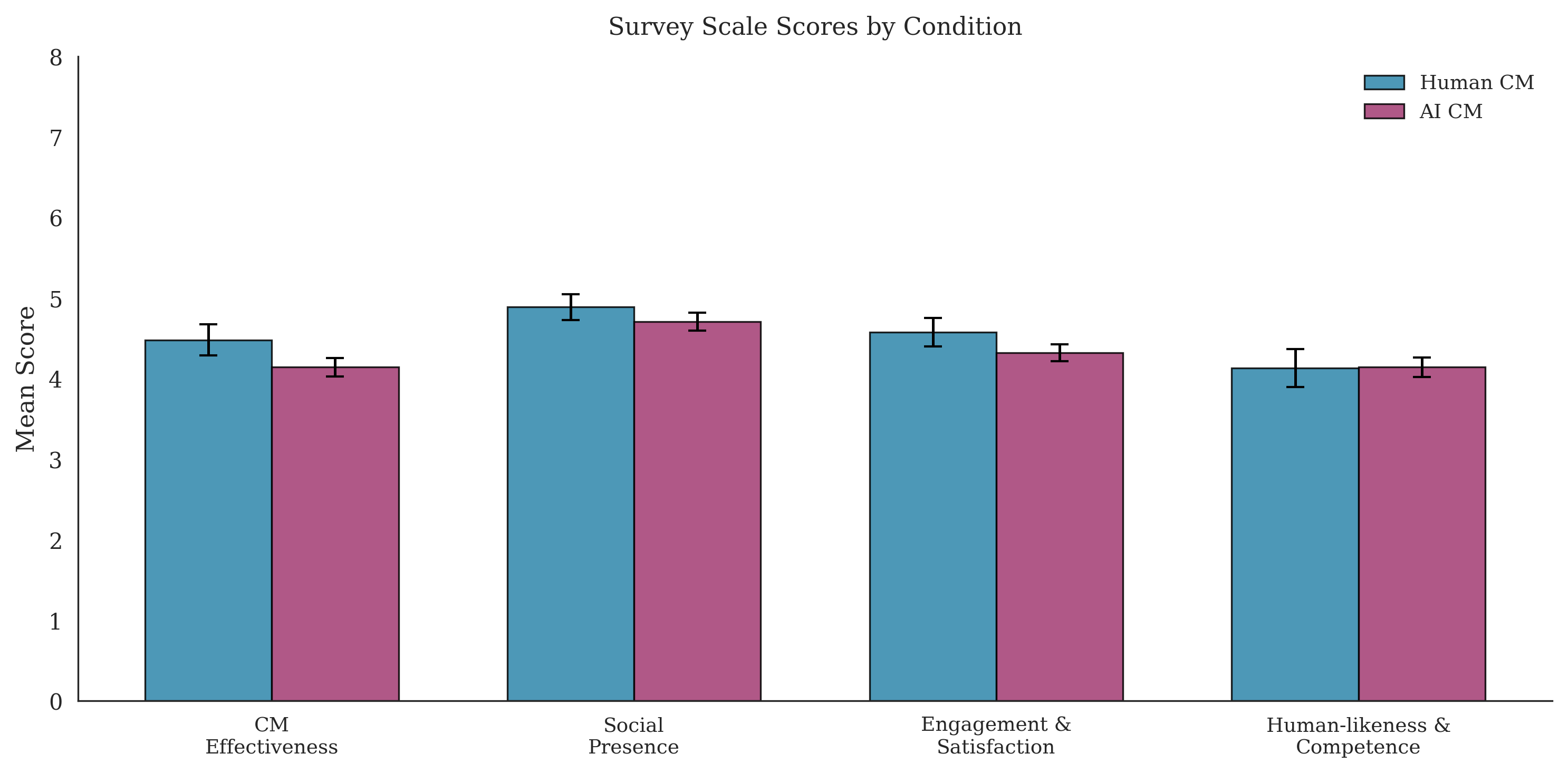}
\caption{Survey scale scores by condition. Bars represent mean scores with standard error of the mean. Differences between conditions were consistently small ($|d| < 0.4$), with substantial overlap in distributions. The pattern of similar ratings across diverse measures indicates comparable participant experiences with human and AI community managers.}
\label{fig:scales}
\end{figure}

All observed differences were small in magnitude ($|d| < 0.4$), with substantial overlap in distributions. Figure~\ref{fig:scales} displays mean scores with standard errors, showing the similarity across conditions. The consistency of small differences across diverse measures suggests that participants had comparable experiences with both types of community managers.

\begin{table*}[t]
\centering
\caption{Survey scale scores by condition. All scales measured on 5-point Likert scales. Differences between conditions were small across all measures.}
\label{tab:scales}
\begin{tabular}{@{}lcccc@{}}
\toprule
\textbf{Scale} & \textbf{Human CM} & \textbf{AI CM} & \textbf{Difference} & \textbf{Cohen's d} \\
 & M (SD) & M (SD) & & \\
\midrule
CM Effectiveness & 4.48 (1.06) & 4.14 (0.84) & 0.34 & -0.37 \\
Social Presence & 4.89 (0.87) & 4.71 (0.83) & 0.18 & -0.21 \\
Engagement \& Satisfaction & 4.58 (0.97) & 4.32 (0.78) & 0.26 & -0.30 \\
Human-likeness \& Competence & 4.13 (1.30) & 4.14 (0.92) & -0.01 & 0.01 \\
\bottomrule
\end{tabular}
\end{table*}

\begin{figure*}[t]
\centering
\includegraphics[width=\textwidth]{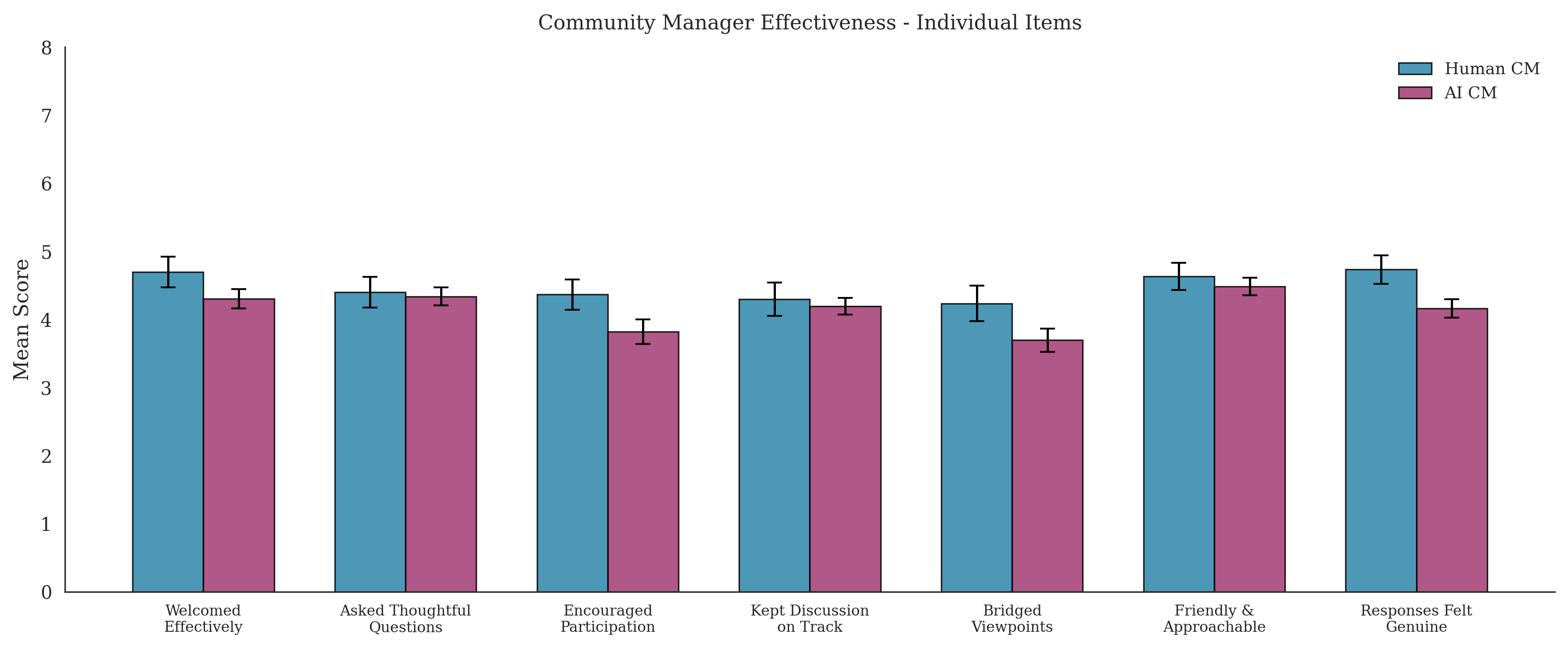}
\caption{Community Manager Effectiveness individual item scores by condition. The scale comprises seven items assessing different aspects of facilitation quality. Bars show mean scores with standard error. While some items (e.g., ``Encouraged Participation,'' ``Bridged Viewpoints'') show modest trends favoring human CMs, differences remain small across all items, consistent with the overall scale similarity.}
\label{fig:cm-items}
\end{figure*}

\subsection{Qualitative Analysis of Detection Explanations}

Participants provided open-ended explanations for their AI detection judgments. Analysis of these responses (n=86 with valid text) revealed attention to diverse cues including response timing, language formality, empathy expression, and conversational flow. However, these cues did not reliably predict actual CM type. Participants citing ``too perfect grammar'' as evidence of AI were equally likely to be evaluating human CMs. Similarly, those noting ``natural conversation flow'' as evidence of human authenticity applied this criterion to both AI and human CMs.

Common reasoning patterns included: response speed or consistency (mentioned by 34\% of participants), language patterns or vocabulary choice (28\%), perceived empathy or emotional understanding (23\%), and conversational naturalness (19\%). These dimensions appear to reflect participants' implicit theories about AI capabilities, but the symmetric error rates demonstrate these theories did not align with observable differences between conditions.

\section{Discussion}
\label{sec:discussion}

Our findings indicate that an AI facilitator designed for natural, asynchronous group chat can achieve near-parity with human facilitators on subjective experience while remaining difficult to identify as nonhuman. This extends prior work on multi-user chatbots by demonstrating the importance of human-like timing and interruption handling alongside strategy selection. HUMA's event-driven design, strategy routing with timeliness regularization, and explicit tool constraints together appear to contribute to believable, restrained participation that avoids dominant or repetitive behavior.

The near-chance detection results suggest that participants' folk theories about AI—centered on speed, grammar, and emotional resonance—may not be diagnostic in this setting. We hypothesize that realistic delays, visible typing indicators, and occasional non-responses (``Keep Silent'') may reduce classic cues participants rely on to spot automation. The small differences in experience measures likely reflect residual artifacts (e.g., consistency of tone or enthusiasm) that future tuning could potentially address.

\subsection{Limitations}

Our study used a single community domain (generative art), short sessions, and role-played personas. Generalization to longer conversations, different cultures, platform affordances, or high-conflict discussions remains to be tested. We did not evaluate long-horizon outcomes (e.g., member retention, knowledge transfer) or measure subtle harms (e.g., over-trust). Finally, while we simulated human timing, we did not model individualized temporal signatures (e.g., personal cadence or diurnal patterns).

\section{Conclusion}
\label{sec:conclusion}

We introduced HUMA, a humanlike multi-user agent framework that addresses the interaction uncanny valley through event-driven architecture, strategic routing with behavioral diversity, realistic timing simulation, and optional participation. Our evaluation demonstrated that AI facilitators implementing these principles achieve near-complete indistinguishability from human facilitators in natural group chat settings, with participants classifying facilitators as human or AI at chance rates while experiencing comparable subjective quality.

\subsection{Limitations and Discussion}

HUMA's success comes with limitations and problems requiring further examination.

\textbf{Use Cases.} The presented use case of community management was well suited for experimental design and transfers into real-world settings. The scope of where HUMA agents could perform well is unknown. Can they be used in work environments? Can they perform together in groups where human participants are a minority?

\textbf{Long-Horizon Social Dynamics.} Our evaluation examined short-term interactions under experimental conditions. Over weeks, months, or years of continuous AI facilitation, do social dynamics evolve with it? Do members form genuine attachment to HUMA agents? Can AI agents support the full lifecycle of community formation, conflict, growth, and decline?

\textbf{Adversarial Robustness and Manipulation.} Human-like AI agents that pass as genuine participants could potentially be weaponized for manipulation, astroturfing, or social engineering at scale. 

% Bibliography
\bibliographystyle{unsrt}
\bibliography{references}

\end{document}